\documentclass[conference, compsocconf]{IEEEtran}

\usepackage{amssymb}
\usepackage[ruled,vlined]{algorithm2e} 
\usepackage[cmex10]{amsmath}
\usepackage{url}
\usepackage{tikz-qtree}
\usepackage{graphicx}
\usepackage{caption}
\usepackage{fixltx2e}
\usepackage{float}
\usepackage{mathtools}
\usepackage[utf8]{vietnam}

\DeclareMathOperator*{\xx}{\mathbf x}

\DeclareMathOperator*{\RR}{\mathbb R}
\DeclareMathOperator*{\EE}{\mathbb E}

\renewcommand{\hat}{\widehat}

\begin{document}

\captionsenglish

\title{
Vietnamese Named Entity Recognition using
Token Regular Expressions and Bidirectional Inference
}

\author{
\IEEEauthorblockN{Phuong Le-Hong}
\IEEEauthorblockA{College of Science\\
Vietnam National University, Hanoi, Vietnam\\
Email: \textit{phuonglh@vnu.edu.vn}}
}

\maketitle

\begin{abstract}
  This paper describes an efficient approach to improve the accuracy
  of a named entity recognition system for Vietnamese. The approach
  combines regular expressions over tokens and a bidirectional
  inference method in a sequence labelling model. The proposed method achieves an
  overall $F_1$ score of $89.66$\% on a test set of an evaluation
  campaign, organized in late 2016 by the Vietnamese Language and
  Speech Processing (VLSP) community.
\end{abstract}

\IEEEpeerreviewmaketitle

\section{Introduction}

Named entity recognition (NER) is a fundamental task in natural
language processing and information extraction. It involves
identifying noun phrases and classifying each of them into a predefined
class. In 1995, the 6th Message Understanding Conference (MUC) 
started evaluating NER systems for English, and in subsequent shared
tasks of CoNLL 2002 and CoNLL 2003 conferences, language independent
NER systems were evaluated. In these evaluation tasks, 
four named entity types were considered, including names of persons,
organizations, locations, and names of miscellaneous entities that do
not belong to these three types.

More recently, the Vietnamese Language and Speech Processing (VLSP)
community has organized an evaluation campaign to systematically
compare NER systems for the Vietnamese language. Similar to the CoNLL
2003 share task, four named entity types are evaluated: persons (PER),
organizations (ORG), locations (LOC), and miscellaneous entities
(MISC). The data are collected from electronic newspapers published on
the web. 

This paper presents the approach and experimental results of our
participating system on this evaluation campaign. In summary, the 
overall $F_1$ score of our system is $89.66$\% on a development set 
extracted from the training dataset provided by the organizing
committee of the evaluation campaign. Three important properties of
our approach include (1) use of token regular expressions to encode
regularities of organization and location names, (2) an algorithm to
annotate every token in an input sentence with their token regular
expression types, and (3) a bidirectional decoding approach to boost
the accuracy of the system. 

The remainder of this paper is structured as
follows. Section~\ref{sec:mlr} gives a brief introduction of
multinomial logistic regression, the main machine learning model which
is used in our system. Section~\ref{sec:features} describes in detail
the features used in our model, including common features used in NER
and those derived from our newly proposed token regular
expressions. This section also presents an algorithm we develop to
annotate every token of an input sentence with its regular expression
type. Section~\ref{sec:decoding} introduces a bidirectional decoding
scheme and a method to combine forward and backward models to get a
better model. Section~\ref{sec:experiments} gives experimental results
and discussions. Finally, Section~\ref{sec:conclusion} concludes the
paper.

\section{Multinomial Logistic Regression}
\label{sec:mlr}

Multinomial logistic regression (a.k.a maximum entropy model) 
is a general purpose discriminative learning method for
classification and prediction which has been successfully applied to
many problems of natural language processing, such as part-of-speech tagging,
syntactic parsing and named entity recognition. In contrast to 
generative classifiers, discriminative classifiers model the posterior
$P(y | \xx)$ directly. One of the main advantages of discriminative
models is that we can integrate many heterogeneous features for
prediction, which are not necessarily independent. Each feature
corresponds to a constraint on the model. In this model, the
conditional probability of a label $y$ given an observation $\xx$ is
defined as
\begin{equation*}
  \label{eq:me}
  P(y|\xx) = \frac{\exp(\theta \cdot 
    \phi(\xx,y))}{\sum_{y \in \mathcal Y}\exp(\theta \cdot 
    \phi(\xx,y))}, 
\end{equation*}
where $\phi(\xx,y) \in \RR^D$ is a real-valued feature
vector, $\mathcal
Y$ is the set of labels and $\theta \in \RR^D$ is the parameter
vector to be estimated from training data. This form of 
distribution corresponds to the maximum entropy probability
distribution satisfying the constraint that the empirical expectation
of each feature is equal to its true expectation in the model:
\begin{equation*}
  \hat \EE(\phi_j(h,t)) = \EE(\phi_j(h,t)), \qquad \forall
  j = 1, 2, \dots, D.
\end{equation*}
The parameter $\theta \in \RR^D$ can be estimated using iterative
scaling algorithms or some more efficient gradient-based optimization
algorithms like conjugate gradient or quasi-Newton
methods~\cite{Andrew:2007}. In this paper, we use the L-BFGS
optimization algorithm~\cite{Nocedal:2006} and $L_2$-regularization
technique to estimate the parameters of the model. This classification model is applied to build
a classifier for the dependency parser where each observation $\xx$ is
a parsing configuration and each label $y$ is a transition type.

\section{Feature Representations}
\label{sec:features}

In discriminative statistical classification models in general and the
maximum entropy model in particular, features play an important role
because they provide the discriminative ability to efficiently
disambiguate classes.  

In order to facilitate the extraction of various feature types, each
lexical token is associated with a surface \textit{word} and an
\textit{annotation map} containing different information of the text
in the form of key and value pairs.  The current annotation map
includes values for part-of-speech, chunk, token regular expression
type and named entity label.

In the context of named entity recognition, the information about
surface word, part-of-speech and chunk tag are given; and in a
training phrase, named entity tags are also provided. In the next
subsection, we describe the regular expression types which are
associated with each token to add some helpful semantic information
for named entity disambiguation.

\subsection{Regular Expressions over Tokens}\label{sec:tokregexp}

We use regular expressions at both character level and token level to
infer useful features for disambiguating named
entities. While character-level regular expressions are used to detect word
shape information, which was shown very important in NER, token-level
regular expressions are very helpful to detect word sequence
information in many long named entities~\cite{Tjong:2003}. 

Common word shape features that our system uses include: 
\begin{itemize}
\item is lower word, e.g., ``\textit{tỉnh}''
\item is capitalized word, e.g., ``\textit{Tổng\_cục}''
\item contains all capitalized letters (allcaps), e.g., ``\textit{UBND}''
\item is mixed case letters, e.g., ``\textit{iPhone}''
\item is capitalized letter with period, e.g., ``\textit{H.}'', ``\textit{Th.}'',
  ``\textit{U.S.}'' 
\item ends in digit, e.g., ``\textit{A9}'', ``\textit{B52}''
\item contains hyphen, e.g., ``\textit{H-P}''
\item is number, e.g., ``\textit{100}''
\item is date, e.g., ``\textit{20-10-1980}'', ``\textit{10/10}''
\item is code, e.g, ``\textit{21B}''
\item is name, where consecutive syllables are capitalized, e.g.,
  ``\textit{Hà\_Nội}'', ``\textit{Buôn\_Mê\_Thuột}'' 
\end{itemize}

Using the word shape features presented above, we then introduce
regular expressions over a sequence of words to capture its
regularity. Suppose that \textit{fPress(w)} is a boolean function
which returns \textit{true} if \textit{w} is in a set of predefined
words related to press and newspaper domain, for example
\textit{\{``báo'', ``tờ'', ``tạp\_chí'', ``đài'',
  ``thông\_tấn\_xã''\}}, and returns \textit{false} otherwise. And
suppose that \textit{fName(w)} is a boolean function which returns
\textit{true} if \textit{w} is a name or an allcaps, and returns
\textit{false} otherwise. Then, we can define the following token
regular expressions to capture the name of a news agency: 
\begin{center}
  \textit{[fPress, fName]}
\end{center}
This sequence pattern matches many different, probably unseen news
agency names, such as \textit{``báo Tuổi\_Trẻ, thông\_tấn\_xã
  Việt\_Nam''}, or \textit{``tờ Batam''}.

In a similar way, suppose that we have a function \textit{fProvince}
which matches common names of administrative structure at various
levels such as \textit{``\{tỉnh, thành\_phố, quận, huyện,
  xã,\dots\}''}, we can build a sequence pattern
\begin{center}
  \textit{[fAllcaps, fProvince, fName]}
\end{center}
which matches many corresponding organization names such as
\textit{``UBND thành\_phố Đà\_Nẵng'', ``HĐND huyện Mù\_Căng\_Chải''},
etc.

Note that an elementary token pattern can be reused in multiple
sequence patterns. For example, the following sequence pattern
\begin{center}
  \textit{[fProvince, fName]}
\end{center}
can match provincial names, which are usually of type location, such
as \textit{``tỉnh Quảng\_Ninh'', ``thành\_phố Hồ\_Chí\_Minh''}.

By examining the training data, we have manually built a dozen of 
regular expresions to match common organization names, and
six regular expressions to match common location names. These regular
expressions over tokens are shown to provide helpful features for
classifying candidate named entities, as shown in the experiments.

\subsection{Regular Expression Type Annotation}

Once regular expressions over tokens have been defined, we add a
regular expression type for each word of an input sentence by
annotating its corresponding annotation map key. Together with word identity, word shape, 
part-of-speech and chunk tag information, regular expression types provide 
helpful information for better classifying named entities, as shown in
the latter experiments. 

We use a greedy algorithm to annotate regular expression type for
every word if it has. Basically, the algorithm works as follows. Given
a sequence of $T$ tokens (or words) $[w_1, w_2, \dots, w_T]$ and a map of regular
expressions over tokens, each key name defines a pattern sequence:
\textit{(patternName, patternRegExp)}, we first search for all
positions of the sentence which begins a pattern match, and select the
longest match, say, pattern \textit{patternName} which ranges from token $w_i$
to token $w_j$, for $1 \leq i < j \leq T$. Then, all the tokens $w_i,
w_{i+1}, \dots, w_j$ are annotated with the same regular expression
type \textit{patternName}. And finally, the algorithm recursively
annotates types for tokens in the remaining two halves of the sequence
$[w_1,w_2,\dots,w_{i-1}]$ and $[w_{j+1},w_{j+2},\dots,w_{T}]$. 

Note that this is a greedy method in that we always choose the longest
pattern in each run. This is a plausible approach since if there are
multiple matches, longer patterns tend to be more correct than shorter
ones. For example, there are two matches on the token sequence
``\textit{UBND tỉnh Đồng\_Nai}'', one is an organization name over
the entire sequence, and another is a location name over the last two
tokens; the longer one is the correct match.

\subsection{Feature Set}

In this subsection, we describe the full feature set that is used in our
system to classify a token at a position of a sentence.

\begin{itemize}
\item Basic features: current word $w_0$, current
  part-of-speech $p_0$, current chunk tag $c_0$,
  previous named entity tags $t_{-1}$ and $t_{-2}$ (or a
  special padding token ``BOS'' -- begin of sentence);
\item Word shape features, as described in the previous subsection;
\item Basic joint features: previous word $w_{-1}$ (or ``BOS''), joint
  of current and previous word $w_0+w_{-1}$, next word
  $w_{+1}$ (or ``EOS'' -- end of sentence), joint of current and
  next word $w_0+w_{+1}$, previous part-of-speech
  $p_{-1}$, joint of current and previous part-of-speech
  $p_0+p_{-1}$, next part-of-speech $p_{+1}$, joint of
  current and next part-of-speech $p_0+p_{+1}$, joint of
  previous and next part-of-speech $p_{-1}+p_{+1}$, joint of
  current word and previous named entity tag $w_0+t_{-1}$;
\item Regular expression types: current regular expression (regexp)
  type $r_0$ (or ``NA'' -- not available), previous regexp
  type $r_{-1}$ (or ``NA''/``BOS''), joint feature
  $r_0+r_{-1}$, next regexp type $r_{+1}$ (or
  ``NA''/``EOS''), joint feature $r_0+r_{+1}$, joint features
  between current word and regexp types $w_0+r_0$,
  $w_0+r_{-1}$, $w_0+r_{+1}$, and lastly, joint features
  between current part-of-speech and regexp types $p_0+r_0$,
  $p_0+r_{-1}$, and $p_0+r_{+1}$.
\end{itemize}

\section{Bidirectional Decoding}
\label{sec:decoding}

The standard decoding algorithm for sequence labelling is the Viterbi
algorithm, which is a dynamic programming algorithm for finding the
most likely sequence of tags given a sequence of observations. In this
work, we also use the Viterbi algorithm to find the best tag sequence
for a given word sequence. However, we found a significant improvement
of tagging accuracy when combining two decoding directions, both
forward decoding and backward decoding. In this section, we describe
our bidirectional decoding approach. 

Given a sequence of $T$ words $[w_1, w_2,\dots,
w_T]$, for each word $w_j$, a pre-trained 
multinomial logistic regression model computes a conditional
probability distribution over possible tags $y_j \in \mathcal Y$: 
\begin{equation*}
  P(y_j|c_j) = \frac{\exp(\theta \cdot \phi(c_j, y_j))}{\sum_{y_j \in
      \mathcal Y} \exp(\theta \cdot \phi(c_j, y_j))}, \quad \forall j =
  1,2,\dots, T, 
\end{equation*}
where $\phi(c_j, y_j)$ is the feature function which extract features from
context $c_j$ containing known information up to
position $j$. As described in the previous section, $c_j$ encodes
useful features for predicting $y_j$, including those extracted from
a local word window $w_{j-2},\dots, w_{j+2}$, two previous tags
$y_{j-1}, y_{j-2}$, and joint features between them. 

The probability of a tag sequence given a word sequence is 
approximated by using the Markov property. In a log scale, we have 
\begin{equation*}
  \log P(y_1,\dots,y_T|w_1,\dots,w_T) \approx \sum_{j=1}^T
  \log P(y_j|c_j).
\end{equation*}

The Viterbi algorithm is then used to find the best tag sequence $\hat y_1,
\hat y_2,\dots, \hat y_T$ corresponding to the max-probability path on a
lattice of size $K\times T$ where $K=|\mathcal Y|$ is the size of the tag set. 

Note that in the second-order Markov model as above, each context
$c_j$ uses the two tags $y_{j-2}$ and $y_{j-1}$ which have been
infered in the previous steps. That said, this is a left-to-right
inference scheme. In the experiments, we use a greedy update
at each position $j$ where the tag $y_j$ is chosen as the best tag of
each local probabilty distribution computed by the maximum entropy
model.

A reversed inference scheme does the same decoding procedure but in a
right-to-left fashion, where two tags $y_{j+2}, y_{j+1}$ are infered
before decoding $y_j$. In essence, when performing backward decoding,
we can use the same Viterbi decoding procedure as in the forward
counterpart, but now using a backward maximum entropy model to compute
the probability of a tag given its following tags. It turns out that
both the training and decoding procedure for this model can be reused simply
by reversing the word and tag sequences at both training and test stages.

An important finding in our experiments is that the backward model is
much better than the forward model in recognizing location names while it
is much worse in recognizing person names. We therefore propose a
method to combine the strength of the two models to boost the accuracy
of the final model. The combination method will be presented in detail
in the experiments.

\section{Experiments}
\label{sec:experiments}

\subsection{Datasets}

We evaluate our system on the training dataset provided by the VLSP NER
campaign.\footnote{\url{http://vlsp.org.vn/evaluation_campaign_NER}}
This dataset contains $16,858$ tagged sentences, totaling $386,520$
words. The dataset contains four different types of named entities: 
person (PER), location (LOC), organization (ORG), and miscellaneous
(MISC). Since the real test set has not been released, we divide this
training set into two parts, one for training and another for
development. The training dataset has $306,512$ tokens (79.3\% of the
corpus), and the development dataset has $80,007$ tokens (20.7\% of
the corpus).

\subsection{Parameter Settings}

The multinomial logistic regression models used in our system are
trained by the L-BFGS optimization algorithm using the 
$L_2$-regularization method with regularization parameter fixed at
$10^{-6}$.\footnote{Using a larger regularization parameter underfits
  the model.} The convergence tolerance of objective function is also
fixed at $10^{-6}$. The maximum number of iterations of the
optimization algorithm is fixed at 300. That is, the training
terminates either when the function value converges or when the number
of iterations is over 300. We use the feature hashing technique as a 
fast and space-efficient method of vectorizing features. The number of
features for our models are fixed at $262,144$ (that is, $2^{18}$).

These
parameters values are chosen according to a series of experiments, for example,
using a smaller number of features (say, $2^{17}$) reduces slightly the performance
of the models, while using a larger number does not result in an improvement of
accuracy but increase the training time.

\subsection{Main Results}

We train our proposed models on the training set and test them on the
development set as described in the previous subsection. The performance of
our system is evaluated on the development set by running the automatic
evaluation script of the CoNLL 2003 shared
task\footnote{\url{http://www.cnts.ua.ac.be/conll2003/ner/}}. The main
results are shown in Table~\ref{tab:1}.

\begin{table}[h]
\center
\caption{Performance of our system}\label{tab:1}
\begin{tabular}{|l|r|r|r|}
  \hline 
  \textbf{Type} & \textbf{Precision} & \textbf{Recall} & \textbf{$F_1$}\\ \hline
  \hline 
  All &$89.56$\%&$89.75$\% & $89.66$ \\ \hline 
  LOC & $84.97$\% &  $94.13$\% & $89.32$ \\ \hline
  MISC & $93.02$\% &  $81.63$\% & $86.96$   \\ \hline
  ORG & $79.75$\% &  $52.72$\% & $63.48$   \\ \hline
  PER & $94.82$\% &  $92.75$\% & $93.77$ \\  \hline
\end{tabular}
\end{table}

Our system achieves an $F_1$ score of 89.66\% overall. Organization
names are the most difficult entity type for the system, 
whose $F_1$ is the lowest of 63.48\%. Person names are the easiest
type for the system whose both precision and recall ratios are high
and the $F_1$ score of this type is 93.77\%. 

\subsection{Effect of Bidirectional Inference}

In this subsection we report and discuss the results using
unidirectional inference, either forward and backward. The performance of
the forward model is shown in Table~\ref{tab:2} and that of the
backward model is shown in Table~\ref{tab:3}. 

\begin{table}[h]
\center
\caption{Performance of the forward model}\label{tab:2}
\begin{tabular}{|l|r|r|r|}
  \hline 
  \textbf{Type} & \textbf{Precision} & \textbf{Recall} & \textbf{$F_1$}\\ \hline
  \hline 
  All &$88.08$\%&$87.10$\% & $87.59$ \\ \hline 
  LOC & $81.61$\% &  $86.54$\% & $84.00$ \\ \hline
  MISC & $97.67$\% &  $85.71$\% & $91.30$  \\ \hline
  ORG & $79.75$\% &  $52.72$\% & $63.48$   \\ \hline
  PER & $94.38$\% &  $93.45$\% & $93.91$  \\  \hline
\end{tabular}
\end{table}

\begin{table}[h]
\center
\caption{Performance of the backward model}\label{tab:3}
\begin{tabular}{|l|r|r|r|}
  \hline 
  \textbf{Type} & \textbf{Precision} & \textbf{Recall} & \textbf{$F_1$}\\ \hline
  \hline 
  All &$88.03$\%&$87.94$\% & $87.98$ \\ \hline 
  LOC & $85.60$\% &  $91.80$\% & $88.59$ \\ \hline
  MISC & $100.00$\% &  $83.67$\% & $91.11$  \\ \hline
  ORG & $66.45$\% &  $43.10$\% & $52.28$   \\ \hline
  PER & $92.15$\% &  $92.54$\% & $92.34$  \\  \hline
\end{tabular}
\end{table}

We see that the backward model is better than the forward model by 4.6 point
of $F_1$ score in recognizing location names. This is surprising
since the only difference between the two models is a reverse of input
sentences. One possible explanation of this effect is that when
recognizing location names of a token sequence $w_1, w_2,\dots,w_n$, if
we already know about the type of $w_n$ it is easier to predict its
previous token $w_{n-1}$ and so on. We conjecture that this is due to the
natural structure of Vietnamese location names.

However, the backward model underperforms the forward model in
recognizing the organization names by a large margin. Its $F_1$ score
of this type is only 52.28\%, while that of the forward model is
63.48\%. This is understandable because our token regular expressions
are designed to capture regularities in many organization names, as
described in the subsection~\ref{sec:tokregexp}, but these expressions
do not work anymore if an input token sequence is reversed.

Either of the two unidirectional models achieves an overall $F_1$ score
of $88.00$\% but when they are combined, our system achieves an overal
score of $89.66$\%, as presented in the previous subsection. The
combined model has both the strong ability of recognizing location
names of the backward model and is good at recognizing organization
names of the forward model.

\subsection{Effect of Token Regular Expressions}

In this subsection, we report the effectiveness of token regular expressions
to our model. We observe that using token regular expressions
significantly improves the performance of the system. 

If the token regular expressions for ORG type are not used, its $F_1$
score of the forward model is $62.94$\%. Adding token regular
experessions for this type help boost this score to
$65.01$\%. Similarly, when token regular expressions for LOC are not
used, its score of the forward model is $82.19$\%. Adding six token
regular expressions for this type improves its score to
$83.07$\%. However, we observe that when all the regular expressions
for this two named entity types are used together, they interact with
each other and make their scores slightly different, as shown in the
Table~\ref{tab:2}.

\subsection{Software}

The named entity recognition system developed in this work has been
integrated into the Vitk toolkit, which includes some fundamental
tools for processing Vietnamese texts. The toolkit is developed in
Java and Scala programming languages, which is open source and
freely downloadable for research
purpose.\footnote{\url{https://github.com/phuonglh/vn.vitk}} An
interesting property of this toolkit is that it is an Apache Spark
application, which is a fast and general engine for large scale data
processing. As a result, Vitk is a very fast and scalable toolkit for
processing big text data. 

\section{Conclusion}
\label{sec:conclusion}

We have introduced our approach and its experimental result in named
entity recognition for Vietnamese text. We have shown the
effectiveness of using token regular expressions, of bidirectional
decoding method in a conditional Markov model for sequence
labelling, and of combining the backward and forward models. Our
system achieves the overall $F_1$ score of $89.66$\% on a test corpus. 

\section*{Acknowledgements}

This research is partly financially supported by Alt
Inc.\footnote{\url{http://alt.ai/corporate}}, and in particular we thank
Dr. Nguyen Tuan Duc, the head of Alt Hanoi office. We thank the
developers of the Apache Spark software.

\bibliographystyle{IEEEtran}
\bibliography{../bibliography}

\end{document}